\documentclass{article}
\usepackage{ijcai19}

\usepackage{times}
\usepackage{soul}
\usepackage{url}
\usepackage[hidelinks]{hyperref}
\usepackage[utf8]{inputenc}
\usepackage[small]{caption}
\usepackage{graphicx}
\usepackage{amsmath}
\usepackage{booktabs}
\usepackage{color}
\usepackage{amsthm, amssymb}
\usepackage{relsize}
\usepackage{subcaption}
\usepackage{caption}

\def\CS{{S}} 
\def\CP{{P}}

\def\mE{\mathop{\mathbb{E}}}

\newcommand{\ff}{f^*\!\!~}

\newcommand{\citet}[1]{\citeauthor{#1} \shortcite{#1}}

\newtheorem{lemma}{Lemma}
\newtheorem{proposition}{Proposition}

\title{Towards Efficient and Unbiased Implementation of Lipschitz Continuity in GANs}

\author{
Zhiming Zhou\and 
Jian Shen\and
Yuxuan Song\and
Weinan Zhang\and
Yong Yu\\
\affiliations
Shanghai Jiao Tong University\\
\emails
heyohai@apex.sjtu.edu.cn
}

\begin{document}

\maketitle

\begin{abstract}
Lipschitz continuity recently becomes popular in generative adversarial networks (GANs). It was observed that the Lipschitz regularized discriminator leads to improved training stability and sample quality. The mainstream implementations of Lipschitz continuity include gradient penalty and spectral normalization. In this paper, we demonstrate that gradient penalty introduces undesired bias, while spectral normalization might be over restrictive. We accordingly propose a new method which is efficient and unbiased. Our experiments verify our analysis and show that the proposed method is able to achieve successful training in various situations where gradient penalty and spectral normalization fail. 
\end{abstract}

\section{Introduction}
Generative adversarial networks (GANs) \cite{gan} is one of the most promising generative models, which has achieved great success in various challenging tasks. The basic framework of GANs consists of a generator and a discriminator, which are both parameterized by neural networks. The generator learns to generate samples to fit the target distribution, while the discriminator measures the distance between the generated distribution and the target distribution. Through a minimax game, the adversarially trained discriminator guides the optimization of the generator. 

In the vanilla GAN, the discriminator is formulated to estimate the Jensen-Shannon (JS) divergence between the two distributions. However, the resulting model suffers from numerous training problems, \emph{e.g.}, gradient vanishing and mode collapse. The common understanding \cite{principled_methods} is that these problems stem from the undesirable property of JS divergence, \emph{i.e.}, when two distributions are disjoint, the JS divergence remains constant and thus cannot provide meaningful guidance to the optimization of the generator. According to \cite{principled_methods}, such a case is very common in practice. Wasserstein distance was thus proposed \cite{wgan} as a new objective of GANs, which can provide continuous measure between two distributions. Empirical experiments demonstrated that Wasserstein distance can significantly improve the training stability. 

Wasserstein distance in its primal form is hard to deal with and is thus usually solved in its dual form \cite{largescaleotmap,wgan}. Wasserstein distance in its dual form requires the discriminative function to be $1$-Lipschitz. 
This arises the problem of how to effectively impose the Lipschitz continuity in the discriminative function. 

Initially, Wasserstein GAN enforces Lipschitz continuity via weight clipping, which was later shown to lead to suboptimal solutions \cite{wgangp,wganlp}. The most common practice of imposing Lipschitz continuity would be gradient penalty as introduced in \cite{wgangp}, which is based on the fact that the Lipschitz constant of a function is equivalent to its maximum gradient scale \cite{adler2018banach} and imposes Lipschitz continuity via penalizing the gradients at sampled points. As an alternative method, \cite{sngan} introduced spectral normalization which restricts the maximum singular value of each layer of the neural network and thereby achieves a global restriction on the Lipschitz constant of the neural network. In this paper, we provide further investigations on the implementation of Lipschitz continuity in GANs. 

First, we will theoretically show that restricting the Lipschitz constant in the blending region of real and fake distributions is sufficient to leverage the theoretical benefit of Wasserstein distance\footnote{The gradient from the optimal discriminative function corresponds to the optimal transport. See Proposition~\ref{proposition1}.}, which indicates that spectral normalization that restricts the global Lipschitz constant of the discriminative function might be over restrictive. We  provide empirical evidences that spectral normalization leads to difficulty in solving of optimal discriminative function. 

On the other hand, we demonstrate that the current implementation of gradient penalty actually introduces extra constraints into the optimization problem, which bias the optimal discriminative function such that the theoretical benefit of Wasserstein distance is impaired. Gradient penalty imposes Lipschitz continuity via penalty method. It is worth noticing that penalty method is soft restriction and the resulting Lipschitz constant is usually larger than $1$. Given that the Lipschitz constant is larger than $1$, many sample points which do not hold the maximum gradient might also have gradients that are larger than $1$. It gives rise to the superfluous constraints imposed by gradient penalty. To regularize the Lipschitz constant, it should only penalize the maximum gradient. The penalties introduced on other sample points are superfluous. We study the impact of these superfluous constraints with synthetic experiments and notice that these superfluous constraints indeed harm the optimization and alter the property of the optimal discriminative function in a bad way.  

Based on the analysis, we propose to impose the Lipschitz continuity in the blending region of real and fake samples via regularizing the maximum gradient. Unlike gradient penalty that casually penalizes all sample gradients, we estimate the maximum gradient and only penalize the maximum gradient, which avoids introducing superfluous constraints. In addition, to provide a method for strict implementation of $k$-Lipschitz, we also present an augmented Lagrangian \cite{alm} based method. The augmented Lagrangian is classic replacement of penalty method where an additional Lagrange multiplier term is introduced. Due to the presence of the Lagrange multiplier, it is able to strictly impose the constraint and thus benefits the situations where strict $k$-Lipschitz is required. 

The remainder of this paper is organized as follows. In Section~\ref{sec_preliminry}, we review the background and current implementations of Lipschitz continuity in GANs. In Section~\ref{sec_analysis}, we analyze the properties of Lipschitz continuity in GANs and the behaviors of existing implementations. In Section~\ref{sec_methods}, we accordingly propose our methods that aim at eliminating the potential issues in the existing methods. We empirically study these methods in Section~\ref{sec_exp} and finally conclude this paper in Section~\ref{sec_conclusion}. 

\section{Preliminaries} \label{sec_preliminry}


\subsection{Wasserstein Distance and Lipschitz Continuity}

Given two metric spaces $(X, d_X)$ and $(Y, d_Y)$, a function $f\colon X \to Y$ is said to be $k$-Lipschitz continuous if there exists some constant $k\geq 0$ such that
\begin{equation} \label{eq_lip}
d_Y(f(x_1), f(x_2)) \leq k \cdot d_X(x_1, x_2), \forall \; x_1, x_2 \in X.
\end{equation}
In this paper and in most existing GANs, the metrics $d_X$ and $d_Y$ are by default Euclidean distance which we denote by $\lVert \cdot \rVert$. The smallest constant $k$ is called the (best) Lipschitz constant of $f$ which we denote by $\lVert f \rVert_{L}$. 

The first-order Wasserstein distance $W_1$ between two probability distributions is defined as
\begin{equation}\label{eq_w_primal}
W_1(\CP_r, \CP_g) =  \inf_{\pi \in \Pi(\CP_r,\CP_g)} \, \mE_{(x,y) \sim \pi} \, [d(x, y)],
\end{equation}
where $\Pi(\CP_r, \CP_g)$ denotes the set of all probability measures with marginals $\CP_r$ and $\CP_g$. It can be interpreted as the minimum cost of transporting the distribution $\CP_g$ to the distribution $\CP_r$. 
The Kantorovich-Rubinstein (KR) duality \cite{oldandnew} provides an more efficient way to compute the Wasserstein distance. The duality states that
\begin{equation} \label{eq_w_dual_lip}
\begin{aligned}
W_1(\CP_r, \CP_g) &= \sup_{f} \, \mE_{x \sim \CP_r} \, [f(x)] \, - \mE_{x \sim \CP_g} \, [f(x)],  \, \\
&\emph{s.t.} \, f(x) - f(y) \leq d(x, y), \,\, \forall x, \forall y. 
\end{aligned}
\end{equation}
The constraint in Eq.~\eqref{eq_w_dual_lip} requires $f$ to be Lipschitz continuous with $\lVert f \rVert_{L} \leq 1$. 

Interestingly, we have the following connection between the optimal solutions in the primal form and dual form \cite{wgangp}.  
\begin{proposition} \label{proposition1}
Let $\pi^*$ be the optimal transport plan in Eq.~(\ref{eq_w_primal}) and $x_t=t x + (1-t) y$ with $0 \leq t \leq 1$. If the optimal discriminative function $\ff$ in Eq.~(\ref{eq_w_dual_lip}) is differentiable and $\pi^*(x,x)=0\,$ for all $x$,  then it holds that
\vspace{-2pt}
\begin{equation} \label{lipgrad}
{P_{(x, y)\sim \pi^*}} \left[\nabla_{\!x_t} \ff(x_t) = \frac{y-x}{\lVert y - x \rVert}\right] = 1. 
\end{equation}
\end{proposition}

The Proposition indicates that the gradient from the optimal $\ff$, which we will later refer as discriminative function in context of GANs, follows the optimal transport plan $\pi^*$.



\subsection{Generative Adversarial Networks}

Generative adversarial networks (GANs) performs generative modeling via a game between two competing neural networks. The generative network learns to map the samples in a prior distribution to a target distribution, while the discriminative network is trained to measure the distance between the target distribution and the distribution of generated samples. The generator and discriminator are connected via a minimax game, so as the generator is able to minimize the distance metric between the two distributions estimated by the adversarially trained discriminator \cite{milgrom2002envelope}.  

In the vanilla GAN \cite{gan}, the discriminator is formulated to estimate the Jensen-Shannon (JS) divergence between the two distributions. However, the resulting model suffers from numerous training problems, \emph{e.g.}, gradient vanishing and mode collapse. According to \cite{principled_methods}, with high dimensional real data, the supports of the target distribution and the generated distribution are very likely to have an intersection of zero measure. However, in such cases, the JS divergence remains constant and is not able to provide meaningful guidance to the optimization of the generator. 

Wasserstein distance was thus proposed \cite{wgan} as an new objective of GANs, which can provide continuous measure between two distributions. Given that the discriminator is well-trained, the generator will receive sustained supervision from the discriminator towards minimizing the Wasserstein distance. Formally, their proposed Wasserstein GAN is defined as follows:
\begin{equation}
\min_G \,\, \max_{f \,\in\, \mathcal{F}} \,\, \mE_{x \sim \CP_r} [f(x))] \,\, - \mE_{z \sim \CP_z} [f(G(z)))], 
\end{equation}
where $\CP_r$ is the target data distribution and $\CP_z$ is the prior distribution (of noise), $G$ and $f$ represent the generative and discriminative function respectively, and $\mathcal{F}$ represent the function class of $\lVert f \rVert_{L} \leq 1$. 




\subsection{Implementations of Lipschitz Continuity}

Wasserstein GAN requires the discriminative function to be $1$-Lipschitz. After that, researchers \cite{kodali2017convergence,fedus2017many,sngan} also empirically found that Lipschitz continuity is also useful when combined with other GAN objectives, \emph{e.g.}, the vanilla GAN objective. Recently, such phenomenon is also theoretically explained \cite{convex_duality,zhou2019lipschitz}, \emph{i.e.}, combining Lipschitz continuity with common GAN objective yields an variant distance metric that is also able to provide continuous measure between the real and fake distributions as Wasserstein distance. As it stands, Lipschitz continuity is a promising technique for improving the training of GANs with theoretical guarantee. However, the implementation of Lipschitz continuity remains challenging. 

Quite a few recent works are devoted to investigate the implementation of Lipschitz continuity. The initial attempt in \cite{wgan} regularizes the Lipschitz continuity via \textbf{weight clipping}, \emph{i.e.}, restricting the maximum value of each weight. It was later shown to lead to suboptimal solution \cite{wgangp,wganlp}. 

And the corresponding alternative methods were  thus proposed for imposing the Lipschitz continuity, named \textbf{gradient penalty} ($L_{gp}$) and \textbf{Lipschitz penalty} ($L_{lp}$) respectively. The two methods share the same spirit and achieve Lipschitz continuity via penalizing the gradient at sampled points towards a given target value (which is usually $1$, however not necessary \cite{progressive_growing_gan,adler2018banach}). They are based on the fact that the Lipschitz constant of a function is equivalent to its max gradient scale \cite{adler2018banach}. 

Formally, the two methods introduce the following regularization terms, respectively: 
\begin{align}
L_{gp} &= -\frac{\rho}{2} \mE_{x \sim \CP_{\hat{x}}} [(\lVert \nabla_x f(x) \rVert - 1)^2], \\
L_{lp} &= -\frac{\rho}{2} \mE_{x \sim \CP_{\hat{x}}} [(\max{0, \lVert \nabla_x f(x) \rVert - 1})^2].
\end{align}
where $\CP_{\hat{x}}$ denotes the sampling distribution defined by the sample strategy which is typically random linear interpolation between real and fake samples. 

\cite{wganlp} argued that gradient penalty is less reasonable because $1$-Lipschitz does not necessarily implies that the gradient scale at every sample point is $1$. It is also the main reason why they proposed to only penalize gradients whose scale is larger than $1$. 

Apart from those already mentioned, \cite{sngan} provide a new direction for enforcing the Lipschitz continuity, named \textbf{spectral normalization} \cite{yoshida2017spectral}, which is based on another fact that the Lipschitz constant of a linear function, $h(x)=Wx$, is equivalent to the weight matrix's maximum singular value. Given the singular value of a weight matrix is easily attainable, they proposed to divide the weight of each linear layer of a neural network by its maximum singular value, \emph{i.e.,}
\begin{equation}
\bar{W}_{SN}=W/\sigma(W),
\end{equation}
where $\sigma(W)$ denotes the maximum singular value of $W$. As the result, the Lipschitz constant of every linear layer is fixed as $1$. Then if the non-linearity parts (\emph{i.e.}, activation functions) are also Lipschitz continuous (which is true for common activation functions), the resulting model will have an upper bound on the Lipschitz constant. 

It is worth noting that spectral normalization results in a hard global restriction on the Lipschitz constant, while gradient penalty and Lipschitz penalty are soft local regularizations. 

\section{Analysis and Motivations}  \label{sec_analysis}

\subsection{The Local Lipschitz Continuity}

The most common choice of $\CP_{\hat{x}}$ in gradient penalty and Lipschitz penalty is the distribution formed by random linear interpolations between the real and fake samples. Currently, why such as choice is valid is still not clear and people tends to believe that it is only a deleterious practical trick \cite{sngan}. 

Here, we provide a theoretical justification as follows. Let $S_{\hat{x}}$ denote the support of the linear interpolations between real and fake distributions. We have  
\begin{lemma}\label{lemma1}
Imposing the Lipschitz continuity over $S_{\hat{x}}$ is sufficient to maintain the property of Proposition~\ref{proposition1}. 
\end{lemma}

To get such conclusion, we need to delve more deep into the well-known KR duality (Eq.~\eqref{eq_w_dual_lip}). The fact is that the constraint in the dual form of Wasserstein distance can be looser than the one in KR duality. 
Specifically, one sufficient constraint for Wasserstein distance in the dual form is as follows: 
\begin{align} \label{w_dual_new}
W_1(&\CP_r, \CP_g) = \sup_{f} \, \mE_{x \sim \CP_r} \, [f(x)] \, - \mE_{x \sim \CP_g} \, [f(x)],  \, \\
&\emph{s.t.} \, f(x) - f(y) \leq d(x, y), \,\, \forall x \in \CS_r, \forall y \sim \CS_g.  \nonumber
\end{align}
where $\CS_r$ and $\CS_g$ denotes the supports of $\CP_r$ and $\CP_g$. Note the difference with KR duality is that: $x$ and $y$ are now from $\CS_r$ and $\CS_g$, instead of being arbitrary, which means the constraints in Eq.~\eqref{w_dual_new} is a subset of the one in Eq.~\eqref{eq_w_dual_lip}. 

It is worthy noticing that given the constraints in Eq.~\eqref{w_dual_new}, any others constraints in Eq.~\eqref{eq_w_dual_lip} does not affect the final result of $W_1(\CP_r, \CP_g)$, and more importantly, any $\ff$ in Eq.~\eqref{w_dual_new} corresponds to one $\ff$ in Eq.~\eqref{eq_w_dual_lip} with the value of $\ff$ on $\CS_r$ and $\CS_g$ unchanged. Thus, any $\ff$ in Eq.~\eqref{w_dual_new} also holds the following key property of Wasserstein distance \cite{oldandnew,wgangp}: 
\begin{lemma} \label{lemma2}
Let $\pi^*$ be the optimal transport plan in Eq.~(\ref{eq_w_primal}) and $\ff$ be the optimal discriminative function in Eq.~(\ref{w_dual_new}). It holds that 
\vspace{-2pt}
\begin{equation} \label{w_dual_p}
{P_{(x, y)\sim \pi^*}} \left[\ff(x)-\ff(y) = d(x,y)\right] = 1. 
\end{equation}
\end{lemma} 

Note that the Proposition~\ref{proposition1} is based on Eq.~\eqref{w_dual_p} and the Lipschitz continuity of $\ff$. And we can further notice that $\ff$ being local Lipschitz continuity over $S_{\hat{x}}$ is sufficient for the proof of Proposition~\ref{proposition1}. The last thing is that $\ff$ being local Lipschitz continuity over $S_{\hat{x}}$ is also a sufficient condition for the constraint in Eq.~\ref{w_dual_new}. Thus, as long as $\ff$ is local Lipschitz continuity over $S_{\hat{x}}$, the Proposition~\ref{proposition1} holds. 

Note that Lemma~\ref{lemma1} also indicates that for training GANs, restricting the global Lipschitz constant might be unnecessary. We next show that although imposing local Lipschitz continuity is sufficient, the current implementations of local Lipschitz continuity is biased. 

\subsection{The Superfluous Constraints}

Gradient penalty and Lipschitz penalty impose the Lipschitz continuity via penalty method. Penalty method is a soft regularization, where the constraint is usually slightly drifted. 

To be more concrete, we consider the following objective, assuming we can directly optimize the Lipschitz constant $k$:  
\begin{equation} \label{dynamic}
L(k) =  W_1(\CP_r, \CP_g, k) \,  - \,\frac{\rho}{2}  (k-1)^2. 
\end{equation}
where $W_1(\CP_r, \CP_g, k)$ denotes the supremum of Wasserstein distance objective, \emph{i.e.}, $\mE_{x \sim \CP_r} [f(x)] - \mE_{x \sim \CP_g} [f(x)]$, under the restriction that $\lVert f \rVert_{L} \leq k$. 

It is clear that $W_1(\CP_r, \CP_g, k) = k W_1(\CP_r, \CP_g)$. Given that $\CP_r$ and $\CP_g$ is fixed, $W_1(\CP_r, \CP_g)$ is a constant. Therefore, $L(k)$ is quadratic function of $k$ and we have that the optimal $k^*$ is $\frac{W_1(\CP_r, \CP_g)}{\rho}+1$. Note that replacing $(k-1)^2$ with $\max\{0, k-1\}^2$ will result in the same optimal $k^*$. 

From the above, we can see that when $\rho$ is small or the distance between $\CP_r$ and $\CP_g$ is large, the resulting Lipschitz constant can be much larger than $1$. Under these circumstances, both gradient penalty and Lipschitz penalty introduce superfluous constraints. Saying the Lipschitz constant is $100$, sampled points with gradient larger than $1$ but smaller than $100$ are penalized, inadvertently. 

We will see in the experiments that these superfluous constraints alter the optimal discriminative function and damage the property of the gradient received by the generator. \cite{wganlp} noted that Lipschitz penalty has a connection to regularized Wasserstein distance. Unfortunately, regularized Wasserstein distance usually also alters the property of the optimal discriminative function and blurs the $\pi^*$ \cite{largescaleotmap}, which is consistent with our analysis here. 

\section{The Proposed Methods} \label{sec_methods}

Now we present our investigation towards more efficient and unbiased implementation of Lipschitz continuity. Given that the local Lipschitz continuity over the support of the linear interpolations between real and fake distributions ($S_{\hat{x}}$) is sufficient, we would consider only restricting the Lipschitz constant in such region. 

\subsection{Max Gradient Penalty}

Similar as gradient penalty, we can regularize the Lipschitz constant via penalty method. But, to avoid the superfluous constraints, we need to only penalize the maximum gradient in $S_{\hat{x}}$. The resulting regularization is as follows:
\begin{align}
L_{maxgp} = -\frac{\rho}{2} ([\max_{x \sim \CP_{\hat{x}}} \big\lVert \nabla_{\!x} f(x) \big\rVert] - 1)^2, 
\end{align}
Analogy to Lipschitz penalty, we can also extend the penalty term with $\max\{0, \cdot\}$. However, when only regularizing the maximum gradient, it is less necessary. Because it will only take effect when the discriminator is underfitting. 

Practically, we follow \cite{wgangp} and sample $x$ as random linear interpolations of real and fake samples in the parallel minibatch. We can either directly use the maximum of the gradient sampled in a minibatch, or keep a historical buffer of the points with maximum gradients which is update in every iteration and then take the maximum gradient over the buffer. The latter is trying to avoid inaccurate estimation of maximum gradient. We have studied this two in experiments. According to our experiments, the buffer is usually unnecessary. Using the maximum gradient in a minibatch would be good enough. 

\subsection{Augmented Lagrangian}

With the penalty method, the constraint is usually not strictly satisfied. The resulting Lipschitz constant, as discussed around Eq.~\eqref{dynamic}, is floating. In the situation where people would like the constraint to be strictly imposed, the augmented Lagrangian is a classic alternative to penalty method, where the constraint would be strictly imposed. In the circumstances of GANs, strictly imposed the $1$-Lipschitz might benefit to control the variable in the contrast experiments, \emph{e.g.}, when comparing different networks and objectives. Also, if one would like to strictly evaluate the Wasserstein distance, a strict restrict of the Lipschitz constant would be favorable. 

The augmented Lagrangian method is a classic method for strict constraint satisfaction. It extends the penalty method by including an extra Lagrange multiplier term. Given that the augmented Lagrangian is a simple extension and there exists potential benefits. We also investigated the practical performance of augmented Lagrangian in imposing of Lipschitz continuity. The regularization term derived from the augmented Lagrangian can be written as follows:
\begin{equation}
L_{maxal} = \lambda (\max_{x \sim \CP_{\hat{x}}} \big\lVert \nabla_{\!x} f(x) \big\rVert - 1) - \frac{\rho}{2} (\max_{x \sim \CP_{\hat{x}}} \big\lVert \nabla_{\!x} f(x) \big\rVert - 1)^2, 
\end{equation}
where $\lambda$ is the Lagrange multiplier. 

The so-called augmented Lagrangian method can also be viewed as an extension of Lagrange multiplier method where only the first term is introduced, and the quadratic penalty term is regarded as the augmentation. Based on the first order optimality of the Lagrange multiplier method and the augmented Lagrangian method, it holds that the optimal $\lambda$ in the Lagrange multiplier method equals to the optimal $\lambda$ in the augmented Lagrangian minus $\rho (\max_{x \sim \CP_{\hat{x}}} \big\lVert \nabla_{\!x} f(x) \big\rVert - 1)$. Thus, there is an common used intuitive update rule for $\lambda$ in augmented Lagrangian, \emph{i.e.}, 
\begin{equation}\label{al_lambda}
\lambda^{k+1} = \lambda^{k} - \rho (\max_{x \sim \CP_{\hat{x}}} \big\lVert \nabla_{\!x} f(x) \big\rVert - 1). 
\end{equation}

Thus, to optimize the augmented Lagrangian, one need only introduce the augmented Lagrangian regularization $L_{maxal}$ and add an extra update step for $\lambda$ according to Eq.~\eqref{al_lambda} after each iteration. 

\section{Experiments} \label{sec_exp}

In this section, we study the practical behaviors of various implementations of Lipschitz continuity, including spectral normalization (SN), gradient penalty (GP), maximum gradient penalty (MAXGP), and the augmented Lagrangian with maximum gradient constraint (MAXAL). In our experiments, Lipschitz penalty shares a very similar performance as gradient penalty. 

We use multilayer perceptron for all toy experiments and use a Resnet architecture \cite{he2016deep} that is similar to the one used in \cite{wgangp} for all other real data experiments. We use Adam optimizer \cite{adam} with $\beta_1=0$, $\beta_2=0.9$. Frechet Inception Distance (FID) \cite{two_time_scale_gan} was used to quantitatively evaluate the resulting models. 
The anonymous code is provided at \url{https://bit.ly/2H4i3Cy}. 

\subsection{Two Dimensional Toy Data}
To intuitively study the property of different methods, we first test their performances with simple two-dimensional data. In this experiment, we randomly sample two data points in two-dimensional space as $\CP_r$ and another two points as $\CP_g$. We fix this two distributions and train a discriminator with different implementations of Lipschitz continuity. 

We want to check whether these methods are able to achieve the optimal discriminative function, by verifying the gradients of generated samples, which should follow the Proposition~\ref{proposition1} and point towards their target real samples that minimize the transport cost. 

\begin{figure}
\centering
\begin{subfigure}{0.325\linewidth}
    \centering
    \includegraphics[width=0.95\columnwidth,height=0.7\columnwidth]{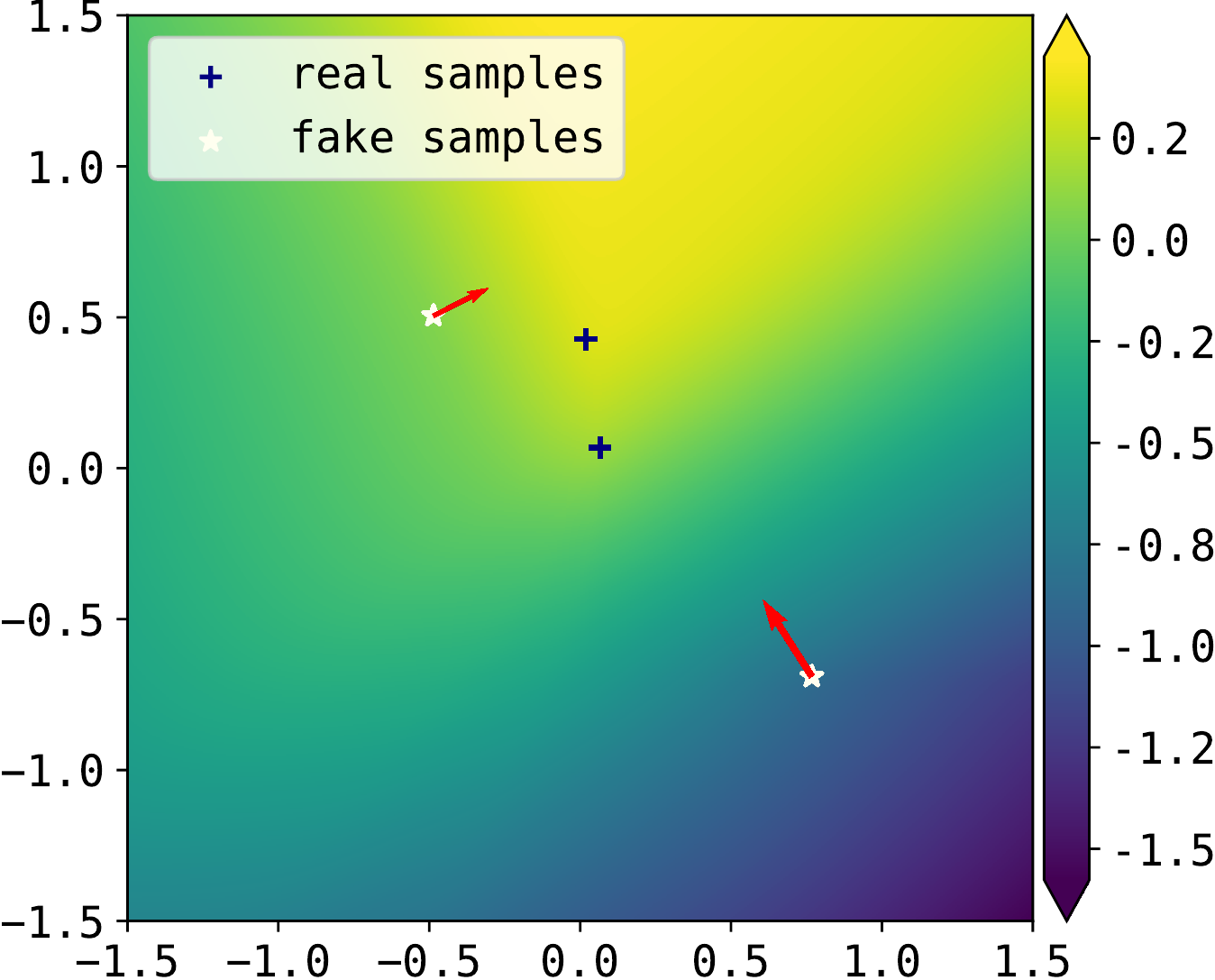}
    \vspace{-3pt}
    \captionsetup{labelformat=empty}
    \caption{SN:1000}
    \label{SN:1000}
\end{subfigure}	
\begin{subfigure}{0.325\linewidth}
    \centering
    \includegraphics[width=0.95\columnwidth,height=0.7\columnwidth]{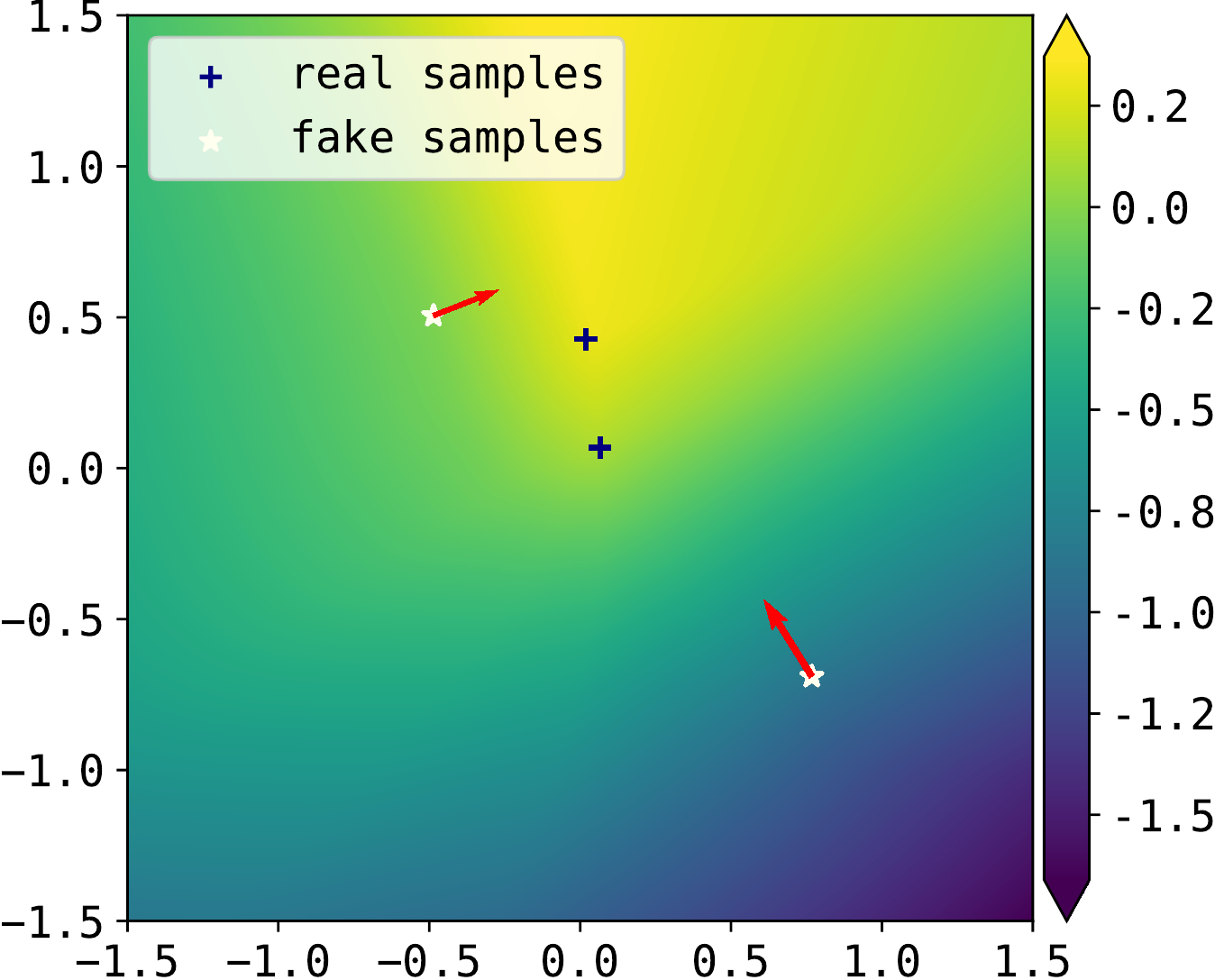}
    \vspace{-3pt}
    \captionsetup{labelformat=empty}
    \caption{SN:2000}
    \label{SN:2000}
\end{subfigure}	
\begin{subfigure}{0.325\linewidth}
    \centering
    \includegraphics[width=0.95\columnwidth,height=0.7\columnwidth]{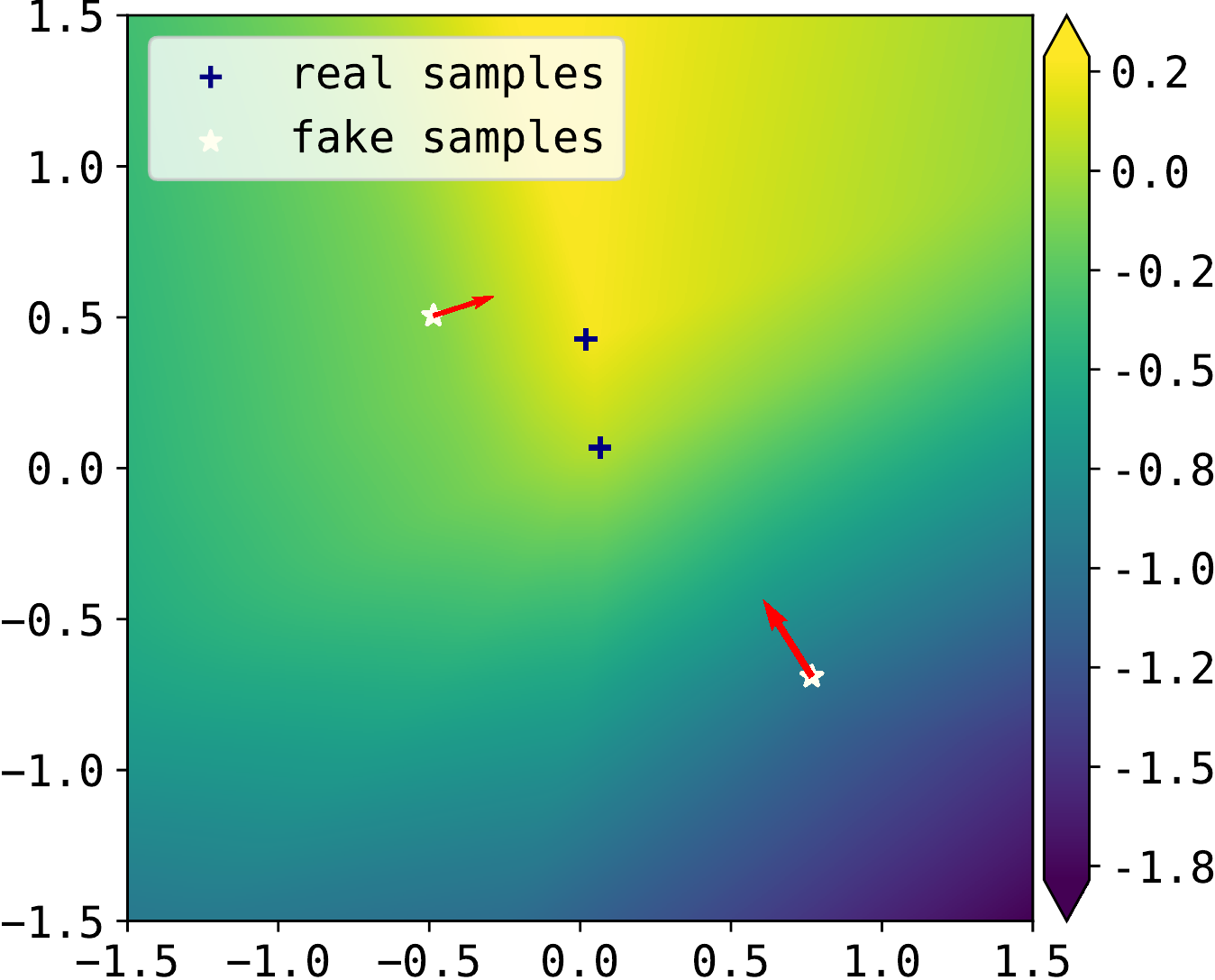}
    \vspace{-3pt}
    \captionsetup{labelformat=empty}
    \caption{SN:3000}
\end{subfigure}	
\begin{subfigure}{0.325\linewidth}
    \centering
    \includegraphics[width=0.95\columnwidth,height=0.7\columnwidth]{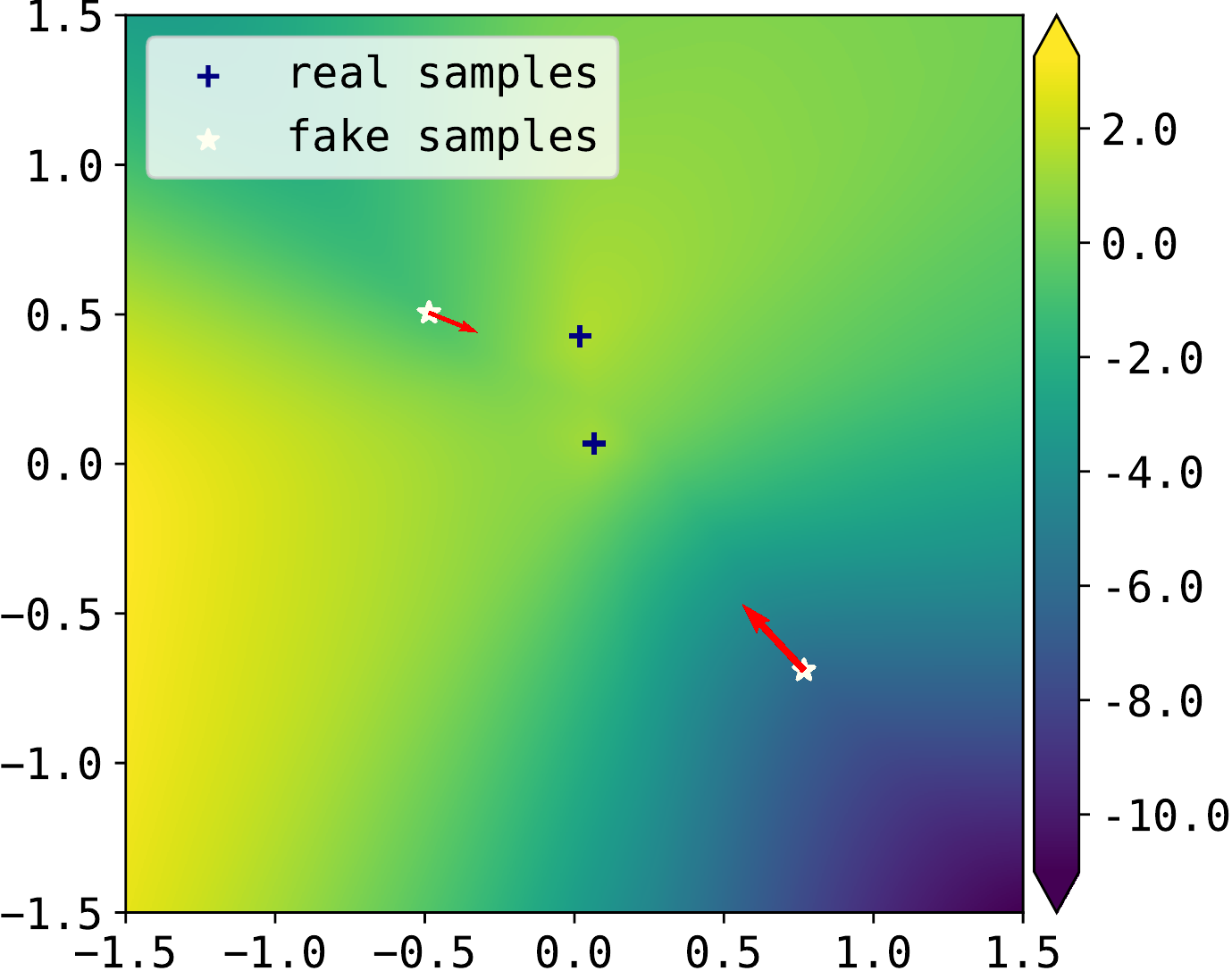}
    \vspace{-3pt}
    \captionsetup{labelformat=empty}
    \caption{GP:1000}
\end{subfigure}	
\begin{subfigure}{0.325\linewidth}
    \centering
    \includegraphics[width=0.95\columnwidth,height=0.7\columnwidth]{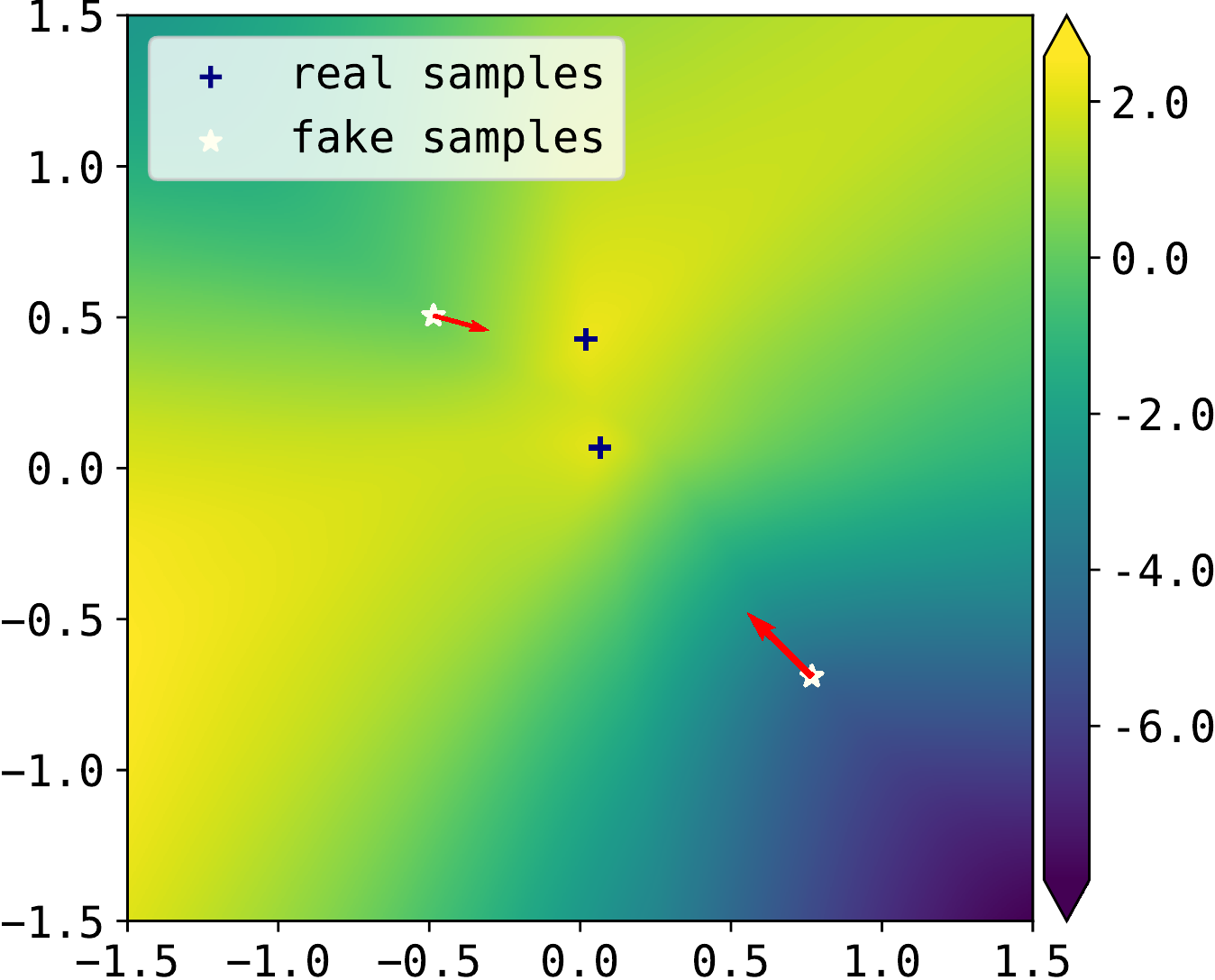}
    \vspace{-3pt}
    \captionsetup{labelformat=empty}
    \caption{GP:2000}
\end{subfigure}	
\begin{subfigure}{0.325\linewidth}
    \centering
    \includegraphics[width=0.95\columnwidth,height=0.7\columnwidth]{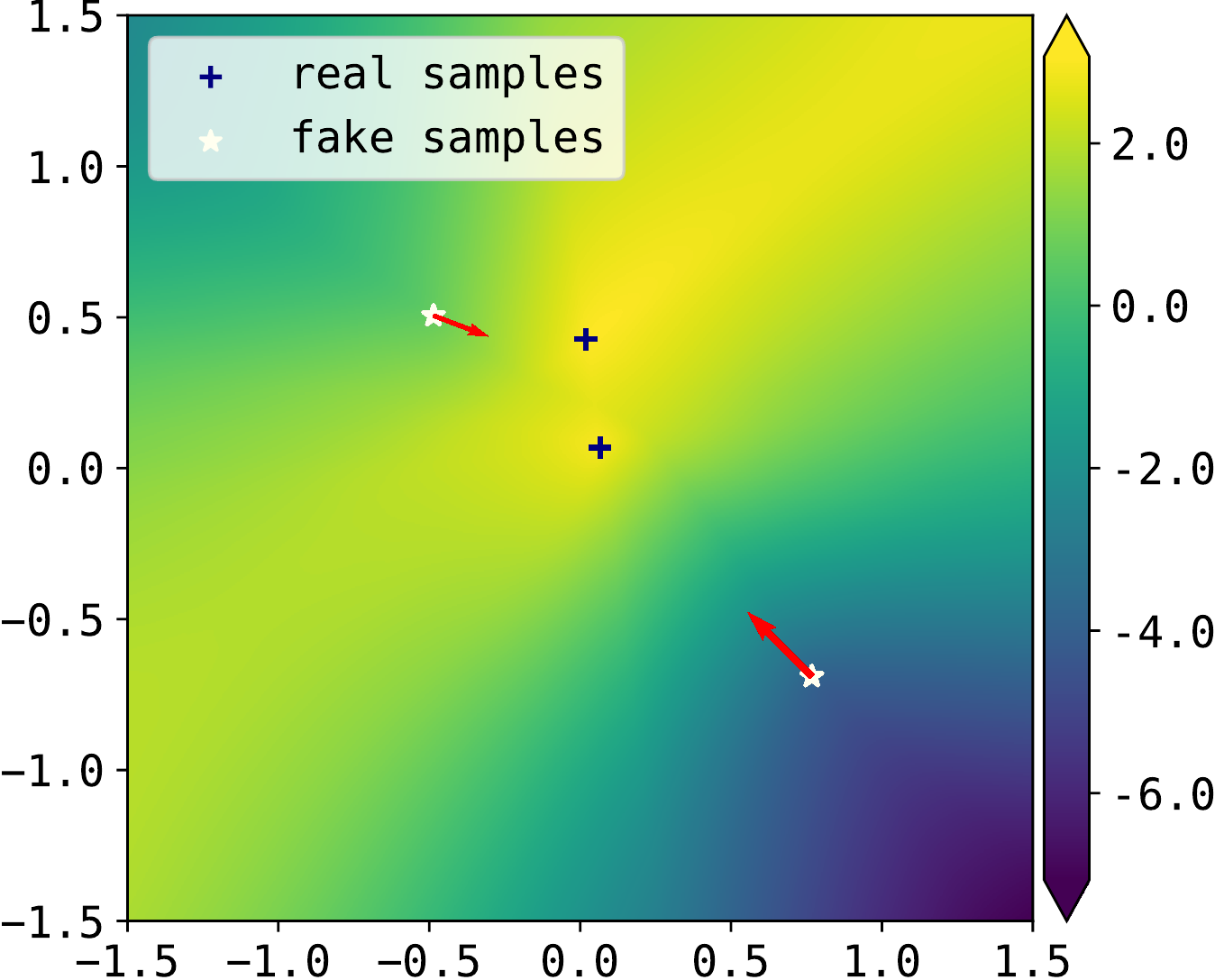}
    \vspace{-3pt}
    \captionsetup{labelformat=empty}
    \caption{GP:3000}
\end{subfigure}	
\begin{subfigure}{0.325\linewidth}
    \centering
    \includegraphics[width=0.95\columnwidth,height=0.7\columnwidth]{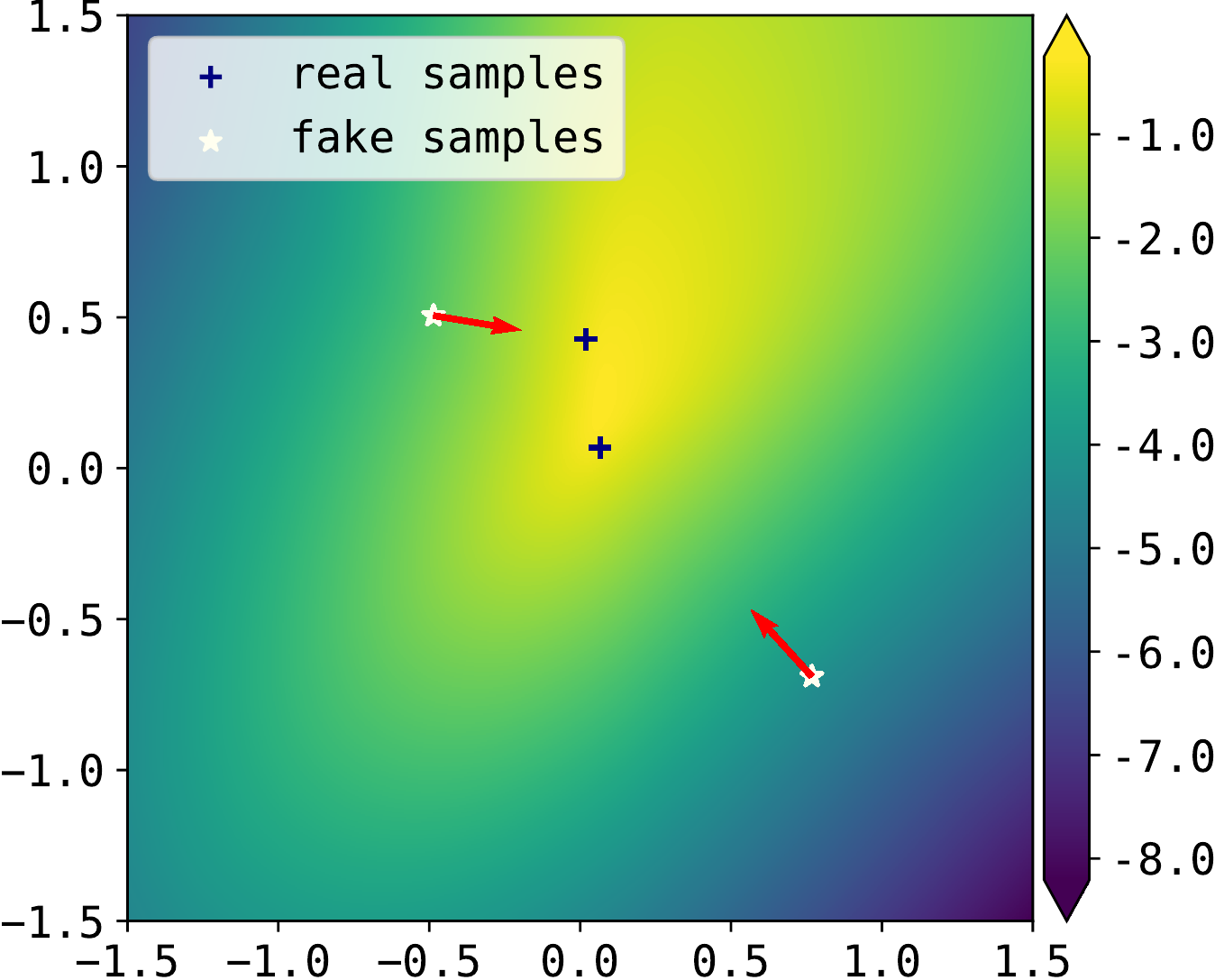}
    \vspace{-3pt}
    \captionsetup{labelformat=empty}
    \caption{MAXGP:1000}
\end{subfigure}	
\begin{subfigure}{0.325\linewidth}
    \centering
    \includegraphics[width=0.95\columnwidth,height=0.7\columnwidth]{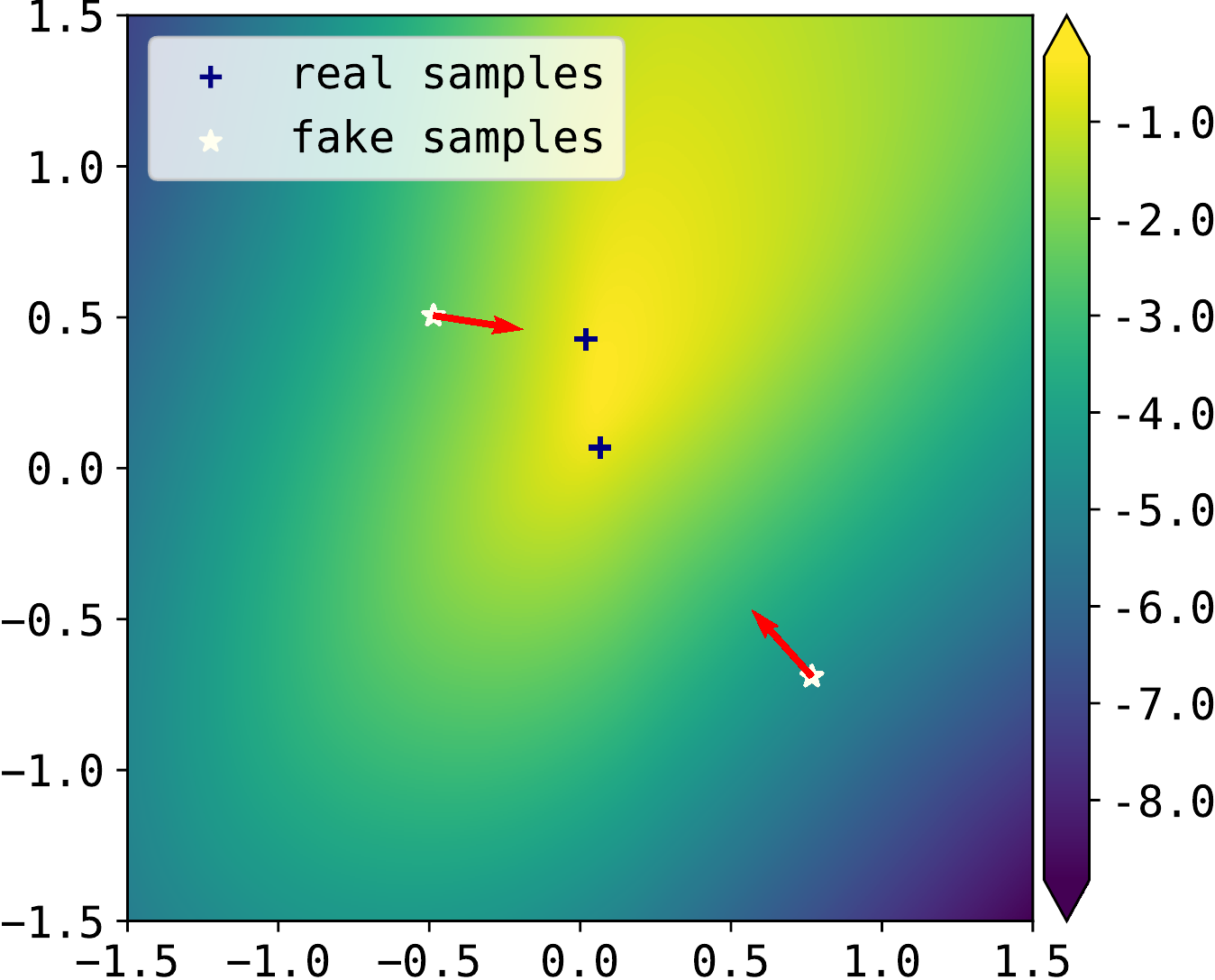}
    \vspace{-3pt}
    \captionsetup{labelformat=empty}
    \caption{MAXGP:2000}
\end{subfigure}	
\begin{subfigure}{0.325\linewidth}
    \centering
    \includegraphics[width=0.95\columnwidth,height=0.7\columnwidth]{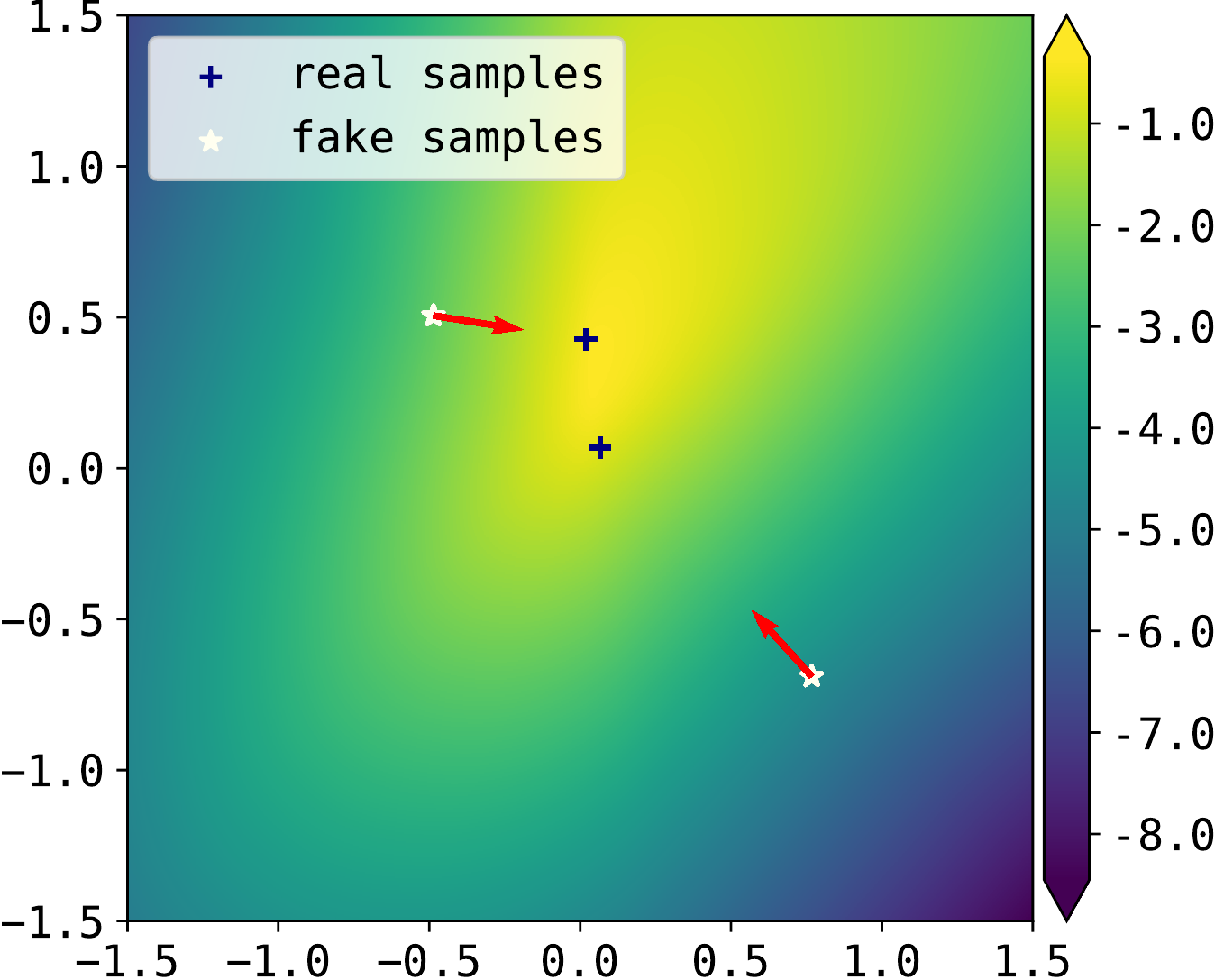}
    \vspace{-3pt}
    \captionsetup{labelformat=empty}
    \caption{MAXGP:3000}
\end{subfigure}	
\vspace{-5pt}
\caption{With $\CP_r$ and $\CP_g$ both being two random sampled points in 2-dimensional space, we training the discriminator using SN, GP and MAXGP, respectively. The number after the name of the methods is the corresponding iteration number. The arrows in the figures indicate the gradient directions. From the results, we notice that: (i) SN in this case failed to achieve the optimal discriminator; (ii) the discriminator trained with GP is oscillatory; (iii) MAXGP stably converged to the optimal.
}
\label{2dtoy}
\end{figure}

Our first interesting observation is that SN in some cases failed to achieve the optimal discriminator. As shown in Figure~\ref{2dtoy}, SN quickly converged to a suboptimal solution and stuck there. We currently do not fully understood how such phenomenon appears. We consider that it might because the global Lipschitz constraint makes the capacity of the discriminator extremely underused such that the optimal discriminative function is not attainable. We have tried fairly large network, but it does not help eliminating this problem. It might also stems from the imperfect singular value estimation of power iteration. We have tried increasing the number of the power iteration that used to acquire the singular value, it does not solve this problem. We have also tried both in-place update of $\hat{W}_{SN}$ and update $\hat{W}_{SN}$ with collection, the problem consistently exists. Training the discriminator for a very long time with decreasing learning rate also cannot solve this problem and the final result keeps unchanged. We would leave further investigation as future work.  

In Figure~\ref{2dtoy}, we also noticed that GP leads to an oscillatory discriminator, which evidences that the superfluous constraints affect the optimal discriminator. By contrast, we see that MAXGP stably converged to the optimal discriminator where the gradients of the fake samples point towards the real samples in an optimal transport way. 

\subsection{Toy Real World Data}

We further compare these methods with real world data. We still want to check whether these method converges to the optimal discriminative function. However, real world dataset is too large, and we found practically, the optimal discriminator is almost non-achievable. Hence, in this experiment, we use a small subset of the real world dataset instead. Specifically, we select ten representative CIFAR-10 images as $\CP_r$ and use ten random noise as $\CP_g$. Then, same as above, we train the discriminator till optimal and check the gradient of the resulting discriminative function of different methods. 

\begin{figure*}
    \vspace{-10pt}
	\centering
	\begin{subfigure}{0.49\linewidth}
		\centering
		\includegraphics[width=0.6\columnwidth]{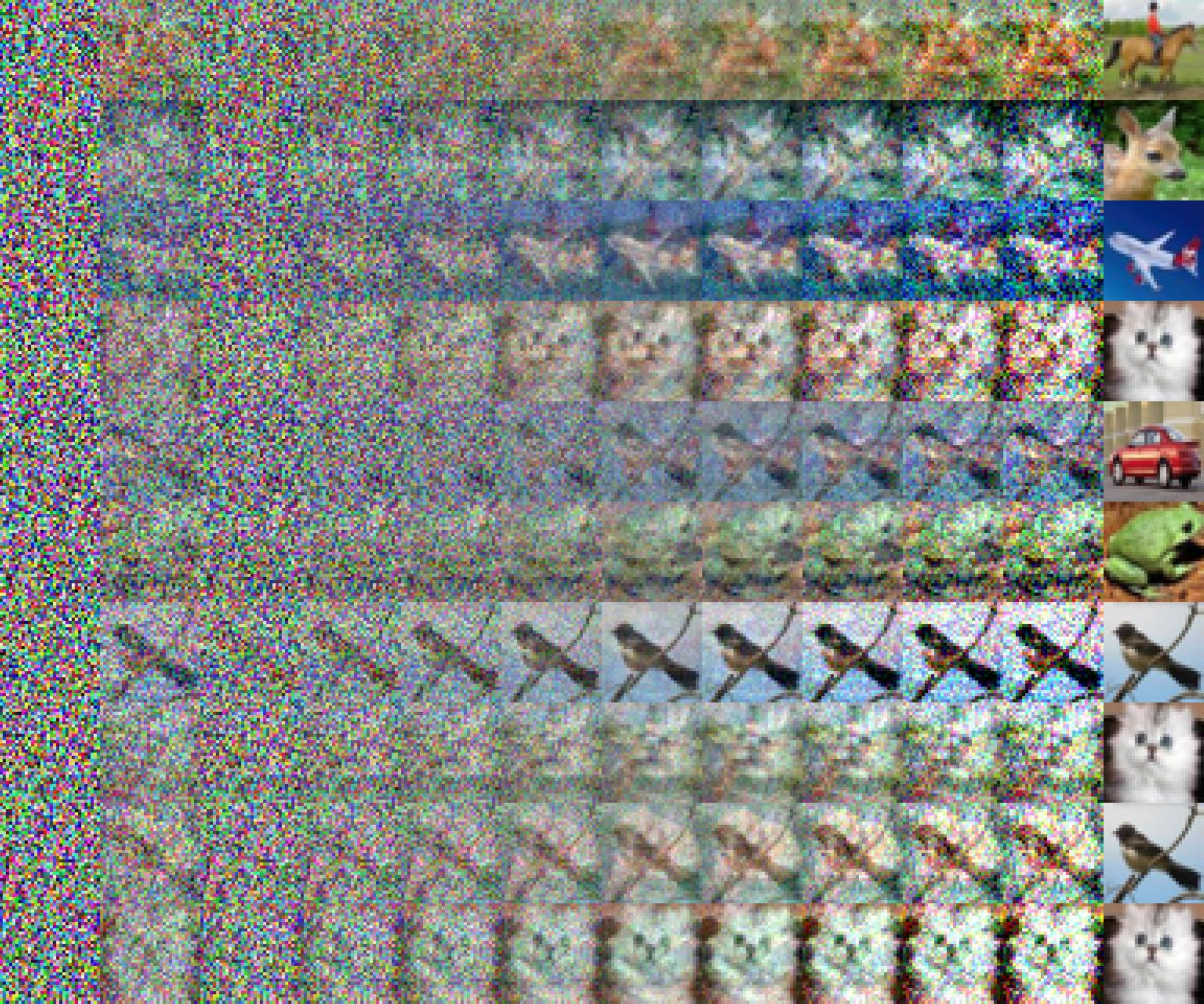}
		\caption{GP}
		\label{fig_gp}
	\end{subfigure}
	\begin{subfigure}{0.49\linewidth}
		\centering	
		\includegraphics[width=0.6\columnwidth]{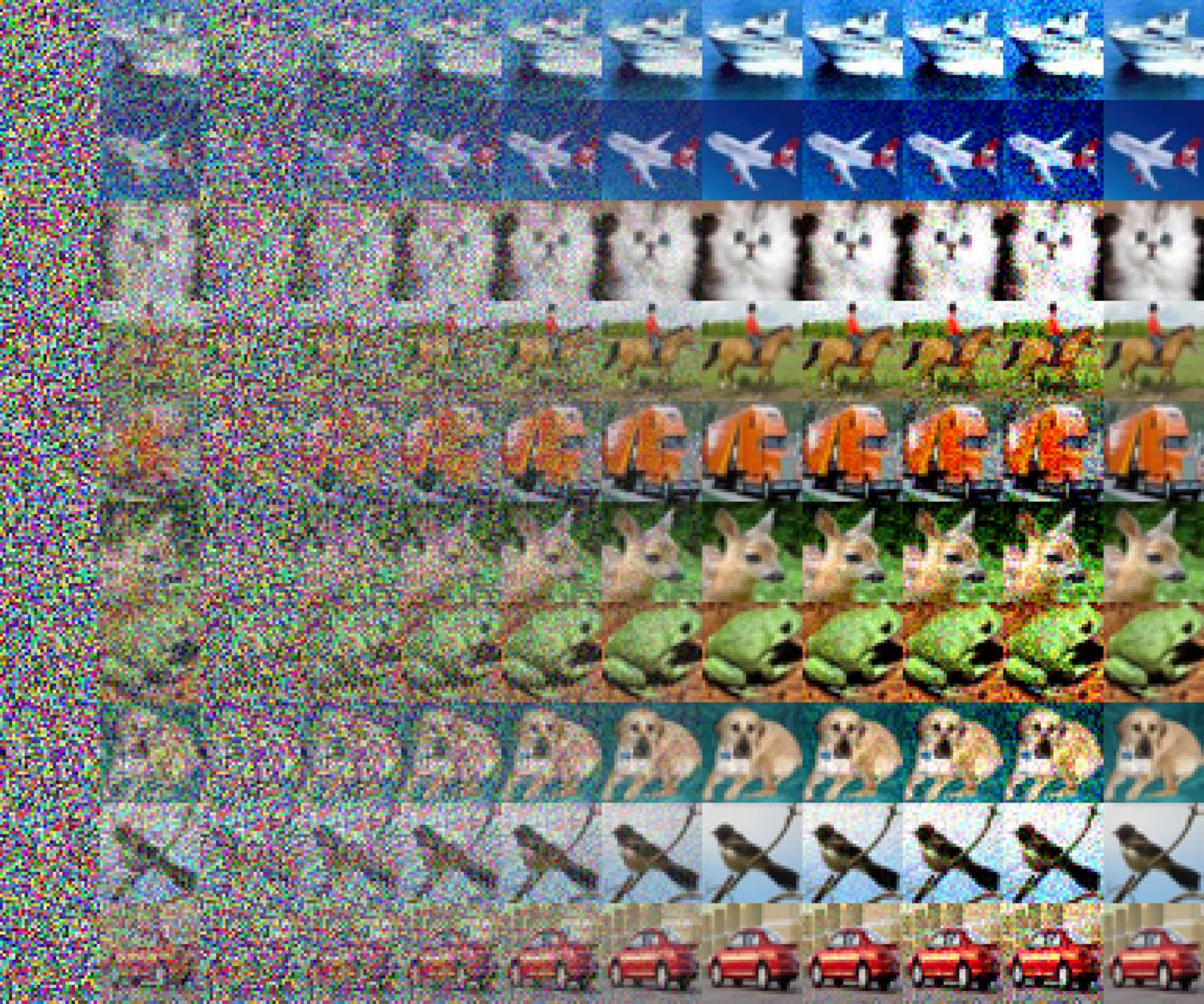}
		\caption{MAXGP}
		\label{fig_maxgp}
	\end{subfigure}	
	\vspace{-5pt}
	\caption{With $\CP_r$ and $\CP_g$ being ten real and noise images, respectively, we train discriminator using GP and MAXGP till optimum. The leftmost in each row is a sample $x$ from $\CP_g$ and the second is the gradient $\nabla_{\!x} f(x)$. The interiors are $x+\epsilon\cdot\nabla_{\!x} f(x)$ with increasing $\epsilon$. The rightmost is the nearest real sample $y$ from $\CP_r$. As we can see, GP failed to achieve the optimal discriminative function, where the gradients of fake samples do not strictly point towards real samples and tend to collapse to a subset of real samples. By contrast, with MAXGP, the gradients of generated samples perfectly follow the optimal transport. } 
	\label{fig_grad}
\end{figure*}

For the high dimensional case, visualizing the gradient direction is nontrivial. Hence, we plot the gradient and corresponding increments. In Figure~\ref{fig_grad}, the leftmost in each row is a sample $x$ from $\CP_g$ and the second is its gradient $\nabla_{\!x} f(x)$. The interiors are $x+\epsilon\cdot\nabla_{\!x} f(x)$ with increasing $\epsilon$, and the rightmost is the nearest (being closest to any point in the incremental path) real sample $y$ from $\CP_r$. 

From the results, MAXGP is also able to achieve the optimal discriminative function in the high dimensional case. We see that the gradient of ten noises in $\CP_g$ is pointing towards the ten real images in $\CP_r$, respectively. However, the resulting gradients of GP do not clearly point towards real samples. The gradient tends to be a blending of several images in the target domain, and it also appears a sort of mode collapse (multiple cats and birds). This experiment once again verifies that these superfluous constraints introduced by GP are harmful. 

\subsection{Sample Quality on CIFAR-10}

\begin{figure}
    \centering
    \begin{subfigure}{0.49\linewidth}
        \centering
        \includegraphics[width=0.99\columnwidth]{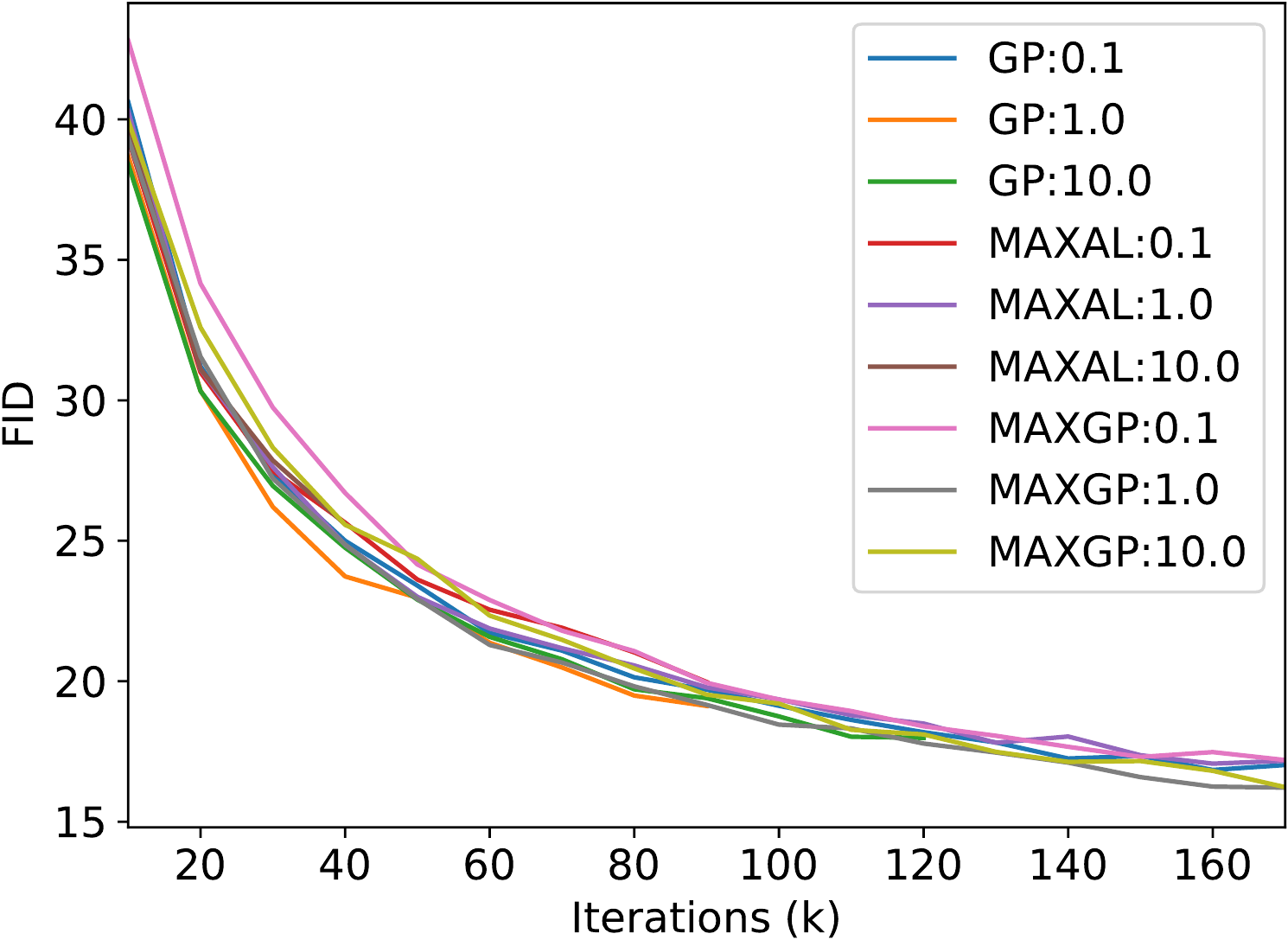}
        \caption{}
        \label{fid_curve_gp}
    \end{subfigure}	
    \begin{subfigure}{0.49\linewidth}
        \centering
        \includegraphics[width=0.99\columnwidth]{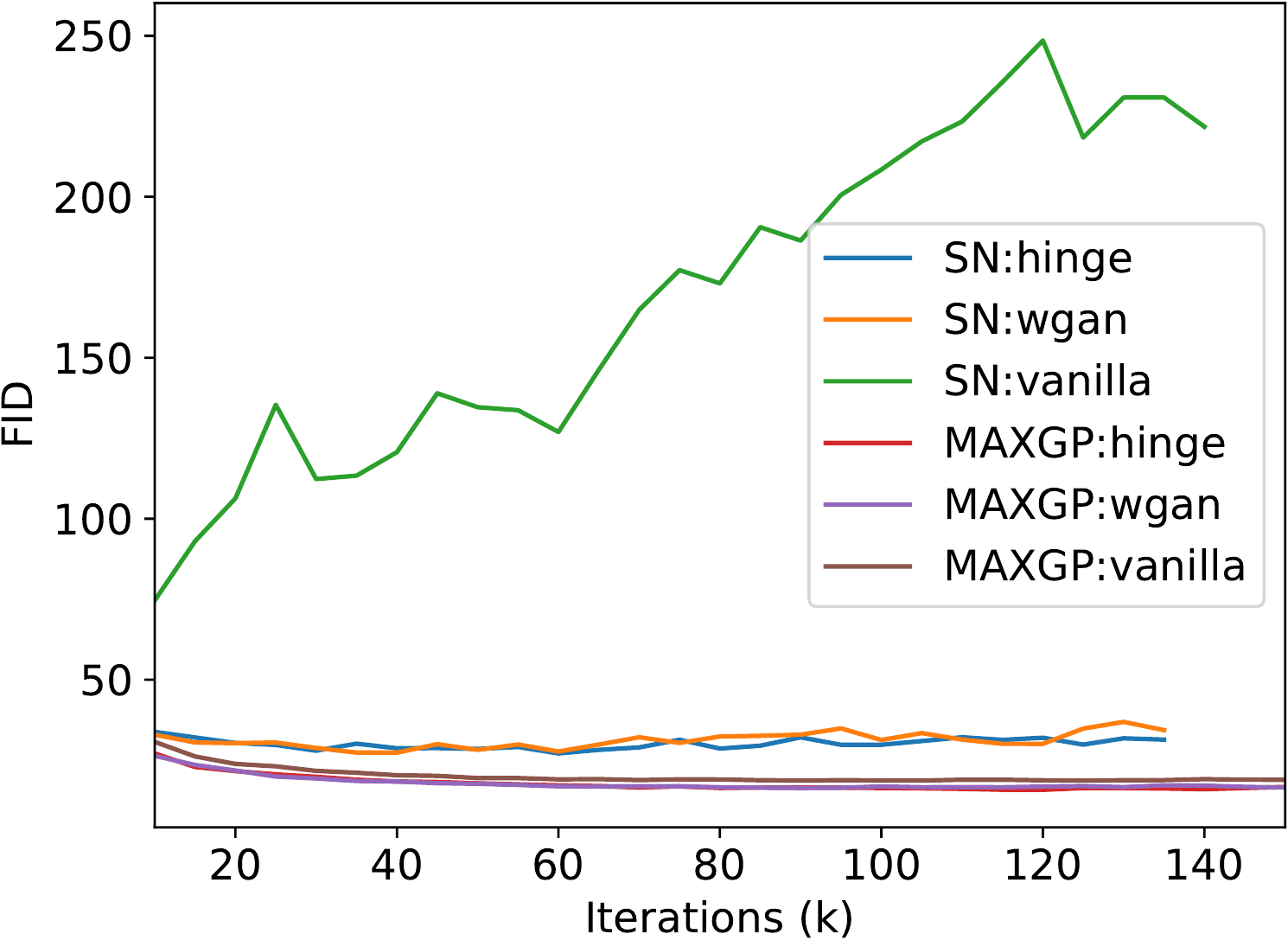}
        \caption{}
        \label{fid_curve_sn}
    \end{subfigure}	
    \vspace{-5pt}
    \caption{Quantitative comparison of unsupervised CIFAR-10 generation in terms of FID training curve. The number after the name of method is the regularization weight $\rho$ and the string after method name indicates the objective it used. GP, MAXGP and MAXAL achieve very similar results and they are not very sensitive to the regularization weight $\rho$. The training of SN diverged when using WGAN objective and even when using the hinge loss or vanilla GAN, the final results are still apparently worse than MAXGP.} 
    \label{fid_curve}
\end{figure}

We now test the practical difference when train a complete GAN model using these methods to impose Lipschitz constraint. In this experiments, we not only train the model with WGAN objective but also with the hinge loss \cite{sngan} and vanilla GAN objective \cite{gan,fedus2017many}, which has also found work well under Lipschitz continuity constraint. The results in terms of training curve of FID are plotted in Figure~\ref{fid_curve}. 

In Figure~\ref{fid_curve_gp}, we compare GP, MAXGP and MAXAL with different regularization weights under the objective of WGAN. We see that the training progresses and final results are quite similar to each other. As we found in the experiments of toy real world data, given $\CP_r$ and $\CP_g$ both consist of ten images, the optimal discriminator is already very hard to achieve. We train the discriminator for $1000k$ iterations with decreasing learning rate to achieve the result in Figure~\ref{fig_grad}. We believe that the reason why these methods do not show obvious difference in these real world applications lies in the optimization level.  That is, in the current hyper-parameter settings, \emph{e.g.}, DCGAN or shallow Resnet, the optimal discriminative function of WGAN is almost impossible to achieve. It might also related to the issues of the optimizer. Amam \cite{adam}, the common-used and somewhat powerful optimizer for GANs, is recently shown to do not guarantee the convergence \cite{reddi2018convergence,zhou2018adashift,zou2018sufficient}.  

In this experiments, we initially use the WGAN objective for all methods. However, we found that with the Resnet architecture \cite{wgangp}, SN failed to converge. The same holds with various small modifications of hyper-parameters. We notice that in \cite{sngan}, when using Resnet architecture, the model with SN is trained using a hinge loss. We therefore also tested SN with the hinge loss, and in additional, the vanilla GAN objective which was also found to also work well given the Lipschitz constraint. The results are plotted in Figure~\ref{fid_curve_sn}. We also included the results of MAXGP with these objectives for comparison. As we can see, the result of MAXGP is generally better than SN. 

\begin{figure}
    \centering
    \begin{subfigure}{0.49\linewidth}
        \centering
        \includegraphics[width=0.99\columnwidth]{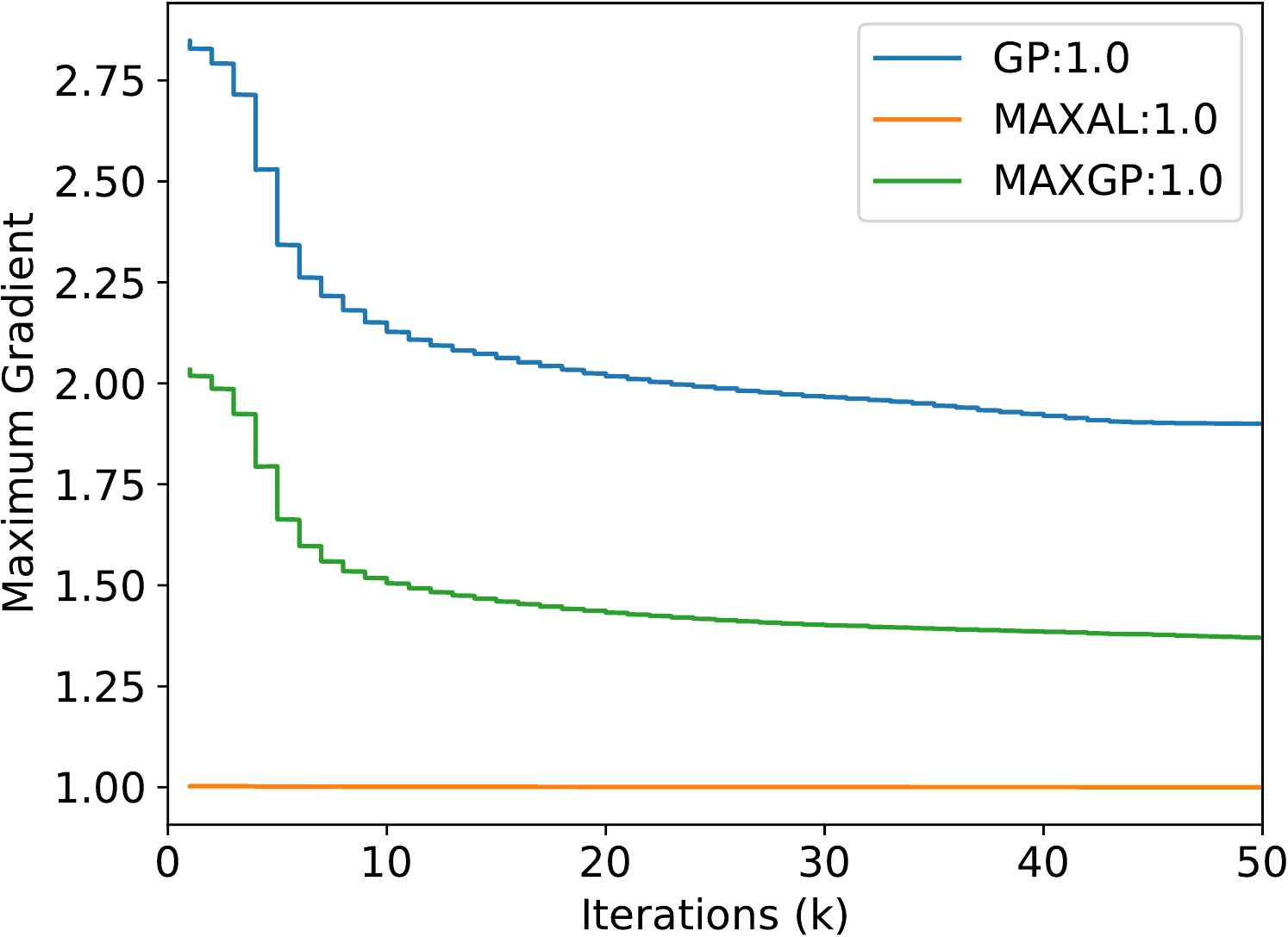}
        \caption{}
        \label{fig_slopes}
    \end{subfigure}	
    \begin{subfigure}{0.49\linewidth}
        \centering
        \includegraphics[width=0.99\columnwidth]{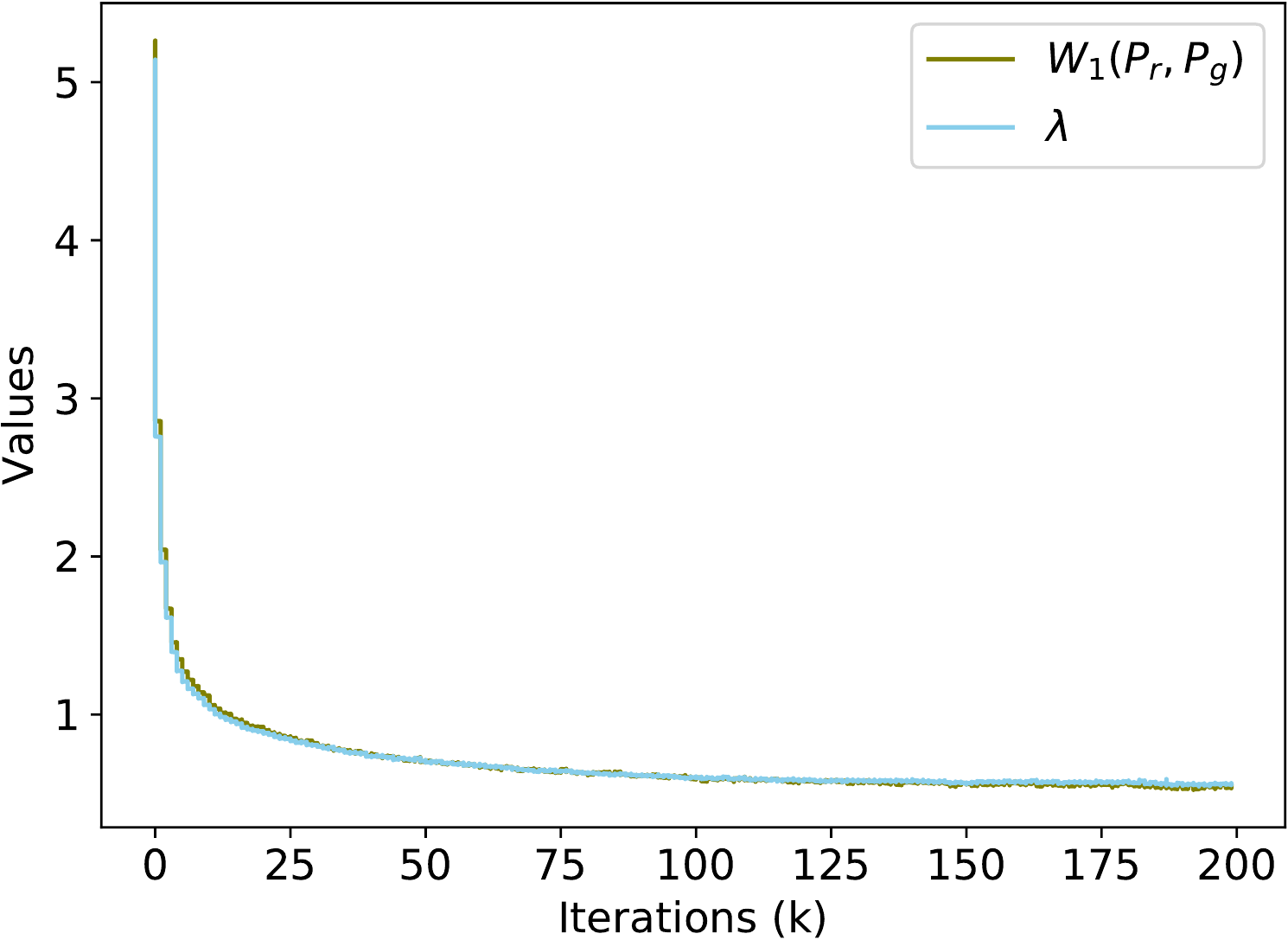}
        \caption{}
        \label{fig_lambda}
    \end{subfigure}	
    \vspace{-5pt}
    \caption{The favorable properties of MAXAL. With MAXAL, the Lipschitz constant quickly converged to the given target. By contrast, the Lipschitz constant of GP and MAXGP is dynamic. In addition, the value of $\lambda$ is equivalent to the Wasserstein distance.}
    \label{fig_maxal}
\end{figure}

Lastly, we inspect the properties of MAXAL. As shown in Figure~\ref{fig_slopes}, MAXAL is able to quickly restrict the Lipschitz constant to the given target $1$ and keep the Lipschitz constant fairly stable during the training. By contrast, the Lipschitz constants under GP and MAXGP keep changing. Another interesting fact about MAXAL is that when trained with the WGAN objective, the optimal $\lambda$ is equivalent to $W_1(\CP_r, \CP_g)$. We verify this fact by plotting this two terms during training together. As showed in Figure~\ref{fig_lambda}, the two lines are basically overlapped. 

\section{Conclusion} \label{sec_conclusion}

In this paper, we demonstrated that restricting the Lipschitz constant over the support of the interpolations of real and fake samples is sufficient to gain the advantageous gradient properties induced by Lipschitz continuity. It provides theoretical guarantee on the validity of these empirical gradient-penalty based methods. In the mean time, it suggests that global restriction on the Lipschitz constant is unnecessary. Combined with the fact that we found the spectral normalization, the method that provides global restriction on the Lipschitz constant somehow fail in many practical scenarios, we suggest to use these methods that regularize local Lipschitz constant. On the other hand, we also observed that the current implementations of local Lipschitz continuity, \emph{i.e.}, gradient penalty and Lipschitz penalty, introduce superfluous constraints to the optimization problem, which evidently alter the optimal discriminative function and impair the favorable gradient properties. 

We have accordingly proposed revision to gradient penalty. Our experiments demonstrated that the proposed method is able to achieve the optimal discriminative function in an unbiased manner. In addition, we suggested augmented Lagrangian as a simple yet good alternative of penalty method which is able to strictly restrict the Lipschitz constant to a given target. 

\bibliographystyle{named}
\bibliography{ijcai19}

\end{document}